\documentclass[10pt, a4paper]{article}

\usepackage[final]{lrec2026} 

\usepackage{graphicx}
\usepackage{amsmath}
\usepackage{amssymb}
\usepackage{booktabs}
\usepackage{tabularx}
\usepackage{array}
\usepackage{makecell}
\usepackage{listings}

\title{\textsc{GROUNDEDKG-RAG}: Grounded Knowledge Graph Index for Long-document Question Answering}

\name{Tianyi Zhang$^{1}$ \quad\quad Andreas Marfurt$^{2}$} 

\address{$^{1}$ getAbstract \quad $^{2}$ Lucerne University of Applied Sciences and Arts\\
         Switzerland \\
         tianyiz0423@gmail.com, andreas.marfurt@hslu.ch\\}

\abstract{
Retrieval-augmented generation (RAG) systems have been widely adopted in contemporary large language models (LLMs) due to their ability to improve generation quality while reducing the required input context length. In this work, we focus on RAG systems for long-document question answering.
Current approaches suffer from a heavy reliance on LLM descriptions resulting in high resource consumption and latency, repetitive content across hierarchical levels, and hallucinations due to no or limited grounding in the source text.
To improve both efficiency and factual accuracy through grounding, we propose \textsc{GroundedKG-RAG}, a RAG system in which the knowledge graph is explicitly extracted from and grounded in the source document. Specifically, we define nodes in \textsc{GroundedKG} as entities and actions, and edges as temporal or semantic relations, with each node and edge grounded in the original sentences. We construct \textsc{GroundedKG} from semantic role labeling (SRL) and abstract meaning representation (AMR) parses and then embed it for retrieval. During querying, we apply the same transformation to the query and retrieve the most relevant sentences from the grounded source text for question answering.
We evaluate \textsc{GroundedKG-RAG} on examples from the NarrativeQA dataset and find that it performs on par with a state-of-the art proprietary long-context model 
at smaller cost and outperforms a competitive baseline. Additionally, our \textsc{GroundedKG} is interpretable and readable by humans, facilitating auditing of results and error analysis.
 \\ \newline  }

\begin{document}

\maketitleabstract

\section{Introduction}


Retrieval-Augmented Generation (RAG) is a technique that retrieves and incorporates additional information from an external knowledge source to answer user queries more accurately. Its application scenarios are diverse. For example, a chatbot may retrieve up-to-date information released after the model’s training cutoff to address users’ questions, or it may search over a confidential contract provided by the user to verify specific terms and conditions.

In this work, we focus on the second scenario, where a long-form document (books with potentially millions of words) is provided, and comprehensive, accurate, and low-latency responses grounded in the document are required.

Current approaches \citep{sarthi2024raptor, edge2025localglobalgraphrag} heavily rely on LLM generations to describe book chunks, entities, relations. They then create a heirarchical tree structure, recursively summarizing clusters of lower-level entities. This reliance on LLM outputs can backfire in two ways.
First, recursive summarization introduces a significant risk of hallucination, as generated summaries may not be strictly grounded in the original text and can introduce content that does not exist in the source document. Second, repeated information across hierarchical levels leads to inefficiencies in both retrieval and prompt construction. Moreover, a fixed tree structure cannot adapt to different queries that may require aggregating different subsets or perspectives of the original document.

To address these issues, we propose \textsc{GroundedKG-RAG}\footnote{The source code is available upon request.}, a RAG system in which every component of the constructed knowledge graph is explicitly grounded in the original text, and aggregation is query-adaptive rather than fixed. We define a \textsc{GroundedKG} whose nodes represent entities and actions, and whose edges encode semantic relations between entities and actions defined by PropBank~\citep{palmer-etal-2005-proposition}, as well as temporal relations between actions. With this clear definition and construction, we only need to embed the nodes while preserving the explicit graph structure. Each node embedding incorporates not only the node itself but also its local graph context including the neighboring nodes.

At query time, we retrieve nodes relevant to the user query and use the graph structure to identify semantically and structurally related nodes. The corresponding grounded text spans are then retrieved and filtered using vector-based semantic similarity and passed to a generative LLM for final answer generation.
The LLM used for answer generation is exchangeable as it does not depend on our retrieval component. We use the same model as our baseline for a fair comparison.

The proposed \textsc{GroundedKG-RAG} framework avoids hallucinations introduced by recursive LLM summarization at multiple hierarchical levels. Every step in the pipeline is human-readable and verifiable due to explicit grounding in the source text. Furthermore, the framework enables efficient and flexible aggregation of document content tailored to different query requirements.

We evaluate \textsc{GroundedKG-RAG} on the NarrativeQA benchmark using Exact Match, Sequence Match, BERTScore~\citep{zhang2020bertscore}, and ROUGE-L F1~\citep{lin-2004-rouge}. Experimental results show that \textsc{GroundedKG-RAG} performs on par with a state-of-the art proprietary long-context model 
at smaller cost, and outperforms the GraphRAG~\citep{edge2025localglobalgraphrag} baseline across all metrics.

\section{Related Work}
Several recent approaches have explored question answering over long documents using retrieval-augmented generation (RAG). One line of work aims to extract node–edge pairs to build a graph index for RAG systems. LightRAG~\citep{guo-etal-2025-lightrag}, HippoRAG~\citep{HippoRAGv1} and HippoRAGv2~\citep{HippoRAGv2} create a knowledge graph from entities and relations in the text. However, these nodes and edges are not clearly defined and contain information at different levels, ranging from words and phrases to paragraphs and passages, making the structure redundant and difficult to organize. Moreover, during retrieval, these methods expand the found nodes with neighboring entities to enable multi-hop reasoning, a common shortcoming of retrieval-augmented generation models \citep{li-etal-2024-retrieval}.

Another line of work relies on LLMs to generate descriptions of document chunks and organize them hierarchically. In RAPTOR~\citep{sarthi2024raptor}, leaf nodes correspond to sentences or fixed-size chunks (e.g., 100 tokens). These nodes are recursively embedded and clustered using semantic similarity, and parent-node summaries are generated by an LLM from their child nodes. These summaries are then used for retrieval. This approach relies heavily on querying LLMs to build summaries, which can lead to redundancy and hallucinations. Another side effect is the use of a fixed tree structure, which may be suboptimal when handling diverse downstream questions.

GraphRAG~\citep{edge2025localglobalgraphrag} integrates knowledge graph construction from the first line of work with the hierarchical structures used in the second approach.
During the indexing stage, it constructs a hierarchical entity-level knowledge graph by prompting large language models (LLMs) to extract entities from the text. These entities are grouped into communities with the Leiden algorithm~\citep{leiden}. The communities are then summarized by LLMs. This process of clustering and summarizing is repeated for a predefined number of levels (1--4 in the paper). In the querying stage, GraphRAG performs a vector search on the embeddings of entities, relations, and communities. It can extend the results with related entities, or even generate new sub questions (in their \textit{drift search}) to expand the search results.

GraphRAG is a compute-intensive and weakly grounded\footnote{Extracted entities, relationships, communities, their descriptions, and importance scores are all generated and evaluated by LLMs without verifiable grounding.} method that relies heavily on LLM generations for its search index. These shortcomings motivated our \textsc{GroundedKG-RAG}, which focuses on grounding the knowledge graph in the source text and is created in a resource-efficient manner.

\section{\textsc{GroundedKG-RAG}}
\subsection{\textsc{GroundedKG} Definition}

To construct a reliable knowledge graph for reasoning that is readable for humans, 
we define a knowledge graph $G$ as a \textit{directed graph}, where nodes represent entities and actions extracted from the unstructured text, and edges represent semantic relations between them. 

Formally, we define the procedure of \textsc{GroundedKG} construction from unstructured text to structured representation as 
$\Phi: \mathcal{T} \rightarrow \mathcal{G}$.
Given a document represented as a sequence of sentences
$S = \{ s_1, \ldots, s_N \} \in \mathcal{T}$,
we construct a grounded knowledge graph
\[
G = \Phi(S) = (V, E),
\]
where $V$ denotes the set of nodes and
$E \subseteq V \times \mathcal{R} \times V$
denotes the set of directed relational edges.

Each node is represented as a dictionary of attributes:
\[
\begin{aligned}
&v: \{\text{node\_id}, \text{label}, \text{texts}, \text{node\_type},\\
&\text{grounded\_texts}\} \longrightarrow \text{Values}.
\end{aligned}
\]

Specifically, for each node $v \in V$,
\[
\begin{aligned}
&v(\text{node\_id}) \in \text{string}, \\
&v(\text{label}) \in \text{string}, \\
&v(\text{texts}) \in \text{list of strings}, \\
&v(\text{node\_type}) \in \{\text{entity}, \text{action}\}, \\
&v(\text{grounded\_texts}) \in \text{list of strings}.
\end{aligned}
\]


Here, \textit{label} is the core concept (e.g., "watch"); \textit{texts} contains textual mentions (e.g., "digital watch"); \textit{grounded\_texts} contains sentences in the document that the node is grounded to.

Similarly, each edge is represented as a dictionary of attributes:
\[
\begin{aligned}
&e: \{\text{source\_node}, \text{target\_node}, 
\text{edge\_role}, \\ &\text{edge\_type}, \text{grounded\_texts}\} \longrightarrow \text{Values}.
\end{aligned}
\]

Specifically, for each edge $e \in {E}$:
\[
\begin{aligned}
&e(\text{source\_node}) \in {V}, \\
&e(\text{target\_node}) \in {V}, \\
&e(\text{edge\_role}) \in \text{String}, \\
&e(\text{edge\_type}) \in \{\text{action--entity}, \text{action--action}\},  \\
&e(\text{grounded\_texts}) \in \text{List[String]}.
\end{aligned}
\]
where \textit{source\_node} and \textit{target\_node} are the source and target node\_ids of the edge; \textit{edge\_role} is the semantic role of the relation, e.g. A0; \textit{grounded\_texts} contains sentences in the document that the node is grounded to.

\subsection{\textsc{GroundedKG} Construction}

\paragraph{\textsc{GroundedKG} from SRL parses.}


Semantic Role Labeling (SRL) is a parsing task that captures the \textit{``who did what to whom, when, where, and how''} structure of a sentence by identifying predicates and assigning semantic roles to their arguments, which describe how each argument participates in an event.

For example, in the sentence \textit{``Peter's mother gave a dose of camomile tea to Peter.''} SRL identifies \textit{gave} as the predicate, \textit{Peter's mother} as the agent (A0), \textit{a dose of camomile tea} as the theme (A1), and \textit{to Peter} as the recipient (A2).

We leverage SRL to parse sentences and construct our \textsc{\textsc{GroundedKG}}. For each predicate, we consider its associated components, including core arguments (A0--A3) as well as temporal and spatial arguments when available.
We first create the graph nodes: each predicate is represented as an \textit{action node}, while all arguments are represented as \textit{entity nodes}.
We then add edges by connecting each predicate to its arguments via \textit{entity--action relations}, for example:
\texttt{(A1, give, a dose of camomile tea, entity--action, <reference to source text>)}.
Finally, for all actions within a sentence, we introduce directed \textit{action--action relations} to encode their temporal order. For the previous example, Peter's mother first makes the tea before giving it to Peter, so the action \textit{give} follows \textit{make}:
\texttt{(next, make, give, action--action, <reference to source text>)}.


\paragraph{\textsc{GroundedKG} from AMR parses.}

Abstract Meaning Representation (AMR) is a semantic parsing task that maps a natural language sentence to a rooted, directed acyclic graph encoding its core semantic meaning, abstracting away from syntactic variation. In an AMR graph, nodes correspond to concepts, including actions and entities, and edges represent semantic relations between these concepts.

For example, in the sentence \textit{``Peter's mother gave a dose of camomile tea to Peter.''} AMR identifies \textit{gave} as the root action (\texttt{give-01} in PropBank), with directed edges to: 
\begin{itemize}
\addtolength{\itemsep}{-1ex}
\vspace{-1ex}
    \item \textit{mother} (A0) as a noun concept which is a person with relations to another person concept \textit{Peter}, 
    \item \textit{tea} as a noun concept as role A1 with quantifier \textit{dose} and modifier \textit{camomile}, 
    \item \textit{Peter} as the person concept (A2).
\end{itemize}

Similar to SRL graph construction, we consider all the concepts including actions and entities. We first create nodes: each action is represented as an \textit{action node}, while all other associated entities are represented as \textit{entity nodes}, and we add modification words to the node text attribute, for example:
\texttt{``(tea, [camomile tea, a dose of camomile tea], tea, <reference to source text>)''}.
We then link nodes with edges by connecting each action to its associated entities via \textit{entity--action relations}, for example:
\texttt{``(A1, give-01, tea, entity--action, <reference to source text>)''}.
Finally, for all actions within a sentence, we introduce directed \textit{action--action relations} to encode their temporal order, for example:
\texttt{``(next, make-01, give-01, action--action, <reference to source text>)''}.

\subsection{Embedding Techniques}
\label{sec:node_embedding}
After constructing the \textsc{GroundedKG}, we create our search index by embedding each node into a vector representation that captures both its semantic meaning and its context within the graph structure. We explore three variants of node embedding: \emph{basic node embedding}, \emph{average neighbor embedding}, and \emph{attention-based neighbor embedding}.

\paragraph{Basic Node Embedding.}
We use a pretrained embedding model $f_\text{embed}$ to encode the \textit{node name} and \textit{node text} into vector representations. The embedding of a node \( v \) is defined as:

\begin{equation}
\begin{aligned}
\mathrm{Embed}(v) = \\
\alpha \, f_\text{embed}(v_{\text{name}})
+ (1 - \alpha) \frac{1}{p} \sum_{i=1}^{p} f_\text{embed}(v_{\text{text}}^{(i)}),
\notag
\end{aligned}
\end{equation}

where \( v \) denotes a node in the \textsc{GroundedKG}, \( v_{\text{name}} \in \texttt{str} \) is the node name, \( v_{\text{text}} = \{v_{\text{text}}^{(1)}, \ldots, v_{\text{text}}^{(p)}\} \) is the set of associated textual descriptions, and \( \alpha \in [0,1] \) is a weighting hyperparameter.

\paragraph{Average Neighbor Embedding.}
Beyond the basic node embedding that relies solely on a node’s name and text, we incorporate the average embeddings of its neighboring nodes to capture local graph context. The resulting node embedding is computed as:

\begin{equation}
\begin{aligned}
\mathrm{NeighborEmbed}(v) =\\
\beta \, \mathrm{Embed}(v)
+ (1 - \beta) \frac{1}{q} \sum_{j=1}^{q} \mathrm{Embed}(v_j),
\notag
\end{aligned}
\end{equation}

where \( \{v_1, \ldots, v_q\} \) are the neighboring nodes of \( v \), and \( \beta \in [0,1] \) is a weighting hyperparameter.
This formulation equally weights contextual information from connected nodes, such as the actions performed by an agent or the participants involved in an event.

\paragraph{Attention-Based Neighbor Embedding.}
To account for varying importance among neighboring nodes, we further introduce an attention-based neighbor embedding. Instead of uniformly averaging neighbor embeddings, we compute attention weights based on cosine similarity between node embeddings. The attention score between node \( v \) and its neighbor \( v_j \) is defined as:

\begin{align}
\mathrm{Attn}(v, v_j)
&= \mathrm{softmax}_j
\bigl( \mathrm{Embed}(v) \cdot \mathrm{Embed}(v_j) \bigr) \notag\\
\notag
\end{align}


where \( \cdot \) denotes the dot product, and \(v_j \in \{v_1, \ldots, v_q\} \). The final node embedding is then computed as:

\begin{equation}
\begin{aligned}
\mathrm{AttentionEmbed}(v) =\\
\beta \, \mathrm{Embed}(v)
+ (1 - \beta) \sum_{j=1}^{q} \mathrm{Attn}(v, v_j)\, \mathrm{Embed}(v_j).
\notag
\end{aligned}
\end{equation}

This attention-based neighbor embedding highlights the most relevant contextual information according to semantic similarity, such as salient actions performed by an agent or the primary participants involved in an event.

\subsection{Retrieval Techniques}

In the querying stage, our goal is to retrieve a subset of query-relevant content
$S_{\text{selected}} \subseteq S$ with high recall. Specifically, we aim to select
grounded texts that are accurate, non-redundant, and that do not omit relevant
evidence required for answering the query. To this end, we design three retrieval
techniques to extract and filter relevant content.

For all retrieval techniques, we first parse the user query using the same graph
construction procedure $\Phi$ defined above. Each query $Q$ is converted into a
query graph
\[
G_q = \Phi(Q) = (V_q, E_q).
\]
Each query node is then embedded following the respective node embedding method described in Section~\ref{sec:node_embedding}.

\paragraph{Basic Node-Based Text Retrieval.}
For each node $u \in V_q$ in the query graph $G_q$, we compute its cosine similarity
with all nodes $v \in V$ in the previously constructed knowledge graph
$G_{\textsc{GroundedKG}}$. We then retrieve the Top-$K$ most similar nodes
($K = 10$ in our experiments) and extract their grounded texts.

Formally, we define the similarity function as
\begin{equation}
\begin{aligned}
\mathrm{sim}(v,u) =
\cos\bigl(\mathrm{Embed}(v) \cdot \mathrm{Embed}(u)\bigr),\\
\quad v \in V,\; u \in V_q.
\notag
\end{aligned}
\end{equation}

For each query node $u_i \in V_q$, the Top-$K$ retrieval is defined as
\[
\mathrm{TopK}(u_i) =
\operatorname*{arg\,topK}_{v \in V} \mathrm{sim}(v, u_i).
\]

The grounded texts associated with the retrieved nodes are then collected as
\[
S_{\text{selected}}(u_i)
= \bigcup_{v_{ij} \in \mathrm{TopK}(u_i)}
v_{ij}.\text{grounded\_texts}.
\]

Finally, the overall selected text set is defined as
\[
S_{\text{selected}}
= \bigcup_{u_i \in V_q} S_{\text{selected}}(u_i),
\quad S_{\text{selected}} \subseteq S.
\]

\paragraph{Text Filter with Vector Similarity.}
In addition to basic node-based retrieval, we further filter the selected texts
using vector similarity at the text level, as is commonly done in contemporary RAG systems. Specifically, we compute the cosine similarity between the query representation $q$ and each selected text $s_i \in S_{\text{selected}}$, and retain the Top-$K$ texts with similarity scores higher than a threshold $\tau$:

\[
\mathrm{VectorSim}_\tau(q)
=
\left\{
s_i \in S_{\text{selected}}
\;\middle|\;
\mathrm{sim}(q, s_i) \ge \tau
\right\}.
\]

\paragraph{Text Filter with Retrieval Count.}
We additionally introduce a salience-based filtering strategy that measures the
importance of each grounded text by counting how frequently it is selected by
different retrieved nodes. Intuitively, if a grounded text $s$ is associated with
multiple selected nodes, it is more likely to be relevant to the query.

Formally, for a grounded text $s_i \in S_{\text{selected}}$, its salience score is
defined as
\[
\mathrm{RetCount}(s_i)
= \sum_{v \in V_{\text{selected}}}
\mathbb{I}\bigl[s_i \in v.\text{grounded\_texts}\bigr],
\]
where $V_{\text{selected}} = \bigcup_{u \in V_q} \mathrm{TopK}(u)$ and
$\mathbb{I}[\cdot]$ is the indicator function.

\section{Experiments}

\subsection{Dataset}
\paragraph{NarrativeQA.}
NarrativeQA \cite{kočiský2017narrativeqareadingcomprehensionchallenge}
is a large reading comprehension benchmark designed to evaluate deep understanding and reasoning over long texts. It consists of 1,572 full-length stories from books and movie scripts, the split for books are 548 train, 58 validation and 177 test. Annotators wrote 46,765 question–answer pairs (around 30 question-answer pairs for each book) based on human-written summaries rather than on the raw text, which encourages models to perform global, integrative reasoning rather than superficial span extraction. Each query requires synthesizing information distributed throughout entire narratives, necessitating comprehension of characters, events, and their relations across extended context spans. Answers are generally short but often not found as direct spans in the stories, reflecting the task’s emphasis on full-content abstraction.

In this experiment, we select three books from the NarrativeQA training split, covering a diverse range of text lengths: The Tale of Peter Rabbit ($\sim$5,000 words), The Phantom of the Opera ($\sim$90,000 words), and Robinson Crusoe ($\sim$120,000 words).

\subsection{Metrics}
We evaluate model performance using a combination of exact-matching, sequence-level, and semantic similarity metrics, capturing both surface-form accuracy and semantic equivalence between predicted and reference answers.

\textbf{Exact Match (EM).}
EM measures whether the predictions exactly contains the reference answers after standard normalization (cleaning up whitespace) and lowercasing. This metric is strict and assigns full credit only when the predicted answer contains an identical segment of the ground truth.

\textbf{Sequence Match (SM).}
SM evaluates whether the predicted answer contains the reference sequence at the token level. Unlike Exact Match, it assigns full credit when predicted answer (e.g. his shoes and his jacket) contains correct reference token sequences (e.g. his shoes and jacket), thereby providing a softer measure of correctness for answers that are lexically similar but not identical.

\textbf{ROUGE-L.}
ROUGE-L~\citep{lin-2004-rouge} measures the longest common subsequence (LCS) between the predicted and reference answers. It captures both precision and recall of overlapping subsequences without requiring contiguous matches, making it effective for evaluating content preservation in free-form or abstractive generation tasks.

\textbf{BERTScore.}
BERTScore~\citep{zhang2020bertscore} computes semantic similarity between predicted and reference answers by aligning contextualized token embeddings from a pretrained transformer model (e.g., BERT). Unlike lexical overlap metrics, BERTScore accounts for paraphrasing and semantic equivalence, providing a robust measure of meaning-level similarity.

\subsection{Experimental Settings}

\textbf{Optimal Settings.}
The best-performing variant of \textsc{GroundedKG-RAG} uses graph construction based on AMR parses, basic node embedding, $K=10$ most similar nodes, and no additional text filters from vector similarity or retrieval counts.

\textbf{Text Segmentation.}
Documents are segmented using the \href{https://reference.langchain.com/python/langchain_text_splitters/}{LangChain} recursive character text splitter into chunks of 5{,}000 tokens with zero overlap, following paragraph boundaries.
Each chunk is further segmented into sentences using \href{https://spacy.io/}{spaCy} with the \texttt{en\_core\_web\_sm} model.

\textbf{Coreference Resolution.}
Coreference resolution is applied to each sentence using spaCy for named entity detection and \href{https://github.com/shon-otmazgin/fastcoref}{fastcoref} for coreference detection. Entities are grouped and grounded to the original sentence.
Each sentence is stored as a tuple \texttt{(sent\_index, original\_sentence, normalized\_sentence, coref\_modified\_sentence)} after processing.

\textbf{Sentence Parsing.}
Semantic parsing is performed on each coreference-resolved sentence using either a semantic role labeling (SRL) parser or an abstract meaning representation (AMR) parser. For SRL parsing, we use spaCy for verb predicate identification and \texttt{cu-kairos/propbank\_srl\_seq2seq\_t5\_large}
for argument extraction.
For AMR parsing, we use \href{https://amrlib.readthedocs.io/en/latest/api_spacy/}{amrlib} for AMR graph extraction together with \href{https://github.com/goodmami/penman}{Penman} for graph parsing. 

\textbf{Graph Construction.} We employ \href{https://networkx.org/en/}{networkx} library for \textsc{GroundedKG} construction and \href{https://pyvis.readthedocs.io/en/latest/}{pyvis} for storing the graph in JSON and html formats.

\textbf{Embeddings.}
Nodes in the constructed graph are embedded using the lightweight sentence embedding model \texttt{all-MiniLM-L6-v2}. For stability, the vector is normalized after each step, including node embedding, aggregated neighbor embeddings, combined node neighbor embeddings. 

\textbf{Retrieval-Augmented Generation.}
Queries are processed using the same pipeline as documents.
Relevant sentences are retrieved based on graph matching and provided as context to OpenAI models \texttt{gpt-5-2025-08-07} and \texttt{gpt-5-nano-2025-08-07} for question answering. We did not observe a big difference in the two model's answer qualities, and therefore selected gpt-5-nano for our further experiments.

\textbf{Hardware.}
All experiments are conducted using a single NVIDIA T4 or L4 GPU accessed on Google Colab, and the OpenAI API with batch processing is used for answer generation.

\section{Results}
\label{sec:results}

\begin{table*}
    \centering
    \small
    \begin{tabular}{ccccc}
    \toprule
Model & Exact match
& Sequence match& Bertscore& RougeL F1\\ \midrule
 \multicolumn{5}{c}{Peter Rabbit: 300 nodes, 700 edges}\\\cmidrule(lr){1-5}
 No context
& \textbf{16}
& 16
& 49
& 22
\\
 Full context
& 13
& 16
& 59
& 32
\\
 GraphRAG& 
\textbf{16}& 
20& 
\textbf{63}& 
8
\\
 \textsc{GroundedKG-RAG}
& \textbf{16}
& \textbf{23}
& 62
& \textbf{34}
\\

\midrule
\multicolumn{5}{c}{The Phantom of the Opera: 25k nodes, 56k edges}\\\cmidrule(lr){1-5}
 No context
& 17
& 17
& 49
& 25
\\
 Full context
& \textbf{34}
& \textbf{37}
& \textbf{62}
& \textbf{36}
\\
 \textsc{GroundedKG-RAG}
& 
\textbf{34}
& 
\textbf{37}
& 56
& \textbf{36}
\\

\midrule \multicolumn{5}{c}{Robinson Crusoe: 33k nodes, 80k edges}\\\cmidrule(lr){1-5}
 No context
& 
13
& 13
& 25
& 8
\\
 Full context
& 36
& 36
& \textbf{55}
& \textbf{31}
\\
 \textsc{GroundedKG-RAG}
& \textbf{40}
& \textbf{43}
& 46
& 
30
\\
\bottomrule 
\end{tabular}%
\caption{Main result of \textsc{GroundedKGRAG} compared to baselines on our three selected books.}
\label{tab:result}
\end{table*}

\begin{table*}
    \centering
    \small
    \begin{tabular}{ccccc}
    \toprule
Model & Exact match
& Sequence match& Bertscore& RougeL F1\\ \midrule
 No context
& \textbf{16}
& 16
& 49
& 22
\\
 Full context
& 13
& 16
& 59
& 32
\\
\midrule
 \textsc{GroundedKG-RAG} with SRL and $K = 5$
& 10
& 16
& 52
& 30
\\
 \textsc{GroundedKG-RAG} with SRL, \\$K=5$, and VectorSim ($\tau=0.2$)
& 10
& 13
& 51
& 25
\\
\midrule
 \textsc{GroundedKG-RAG}
& \textbf{16}
& \textbf{23}
& \textbf{62}
& 34
\\
 \textsc{GroundedKG-RAG} with $K = 5$
& 13
& 20
& 58
& 34
\\
 \textsc{GroundedKG-RAG} with VectorSim ($\tau = 0.2$)
& 13
& 16
& 58
& 
33\\
\midrule
 \textsc{GroundedKG-RAG} (basic Embed)
& \textbf{16}
& \textbf{23}
& \textbf{62}
& 34
\\
 NeighborEmbed ($\beta = 0.8$)
& \textbf{16}
& \textbf{23}
& 60
& 33
\\
 AttentionEmbed ($\beta = 0.5$)
& 13
& 20
& 60
& \textbf{36}
\\
 AttentionEmbed ($\beta = 0.8$)
& \textbf{16}
& \textbf{23}
& 60
& \textbf{36}
\\
\bottomrule 
    \end{tabular}
    \caption{Ablation of our \textsc{GroundedKG-RAG} on Peter Rabbit.}
    \label{tab:ablation-pr}
\end{table*}

We evaluate our \textsc{GroundedKG-RAG} by comparing with different baselines and ablating our design choices.

\textbf{Baseline Comparison.}
Our main results are shown in Table~\ref{tab:result}. We compare our \textsc{GroundedKG} with a closed-book LLM (without any context, \textit{no context}), an open-book LLM (offering the full book content as context, \textit{full context}), and GraphRAG. Since the NarrativeQA questions are based on copyright-free books from \href{https://www.gutenberg.org/}{Project Gutenberg}, we assume that both the books and potentially also the NarrativeQA question-answer pairs have been part of the proprietary LLM's training data. The \textit{no context} baseline establishes the performance of the answer generation model when only retrieving (book text or memorized NarrativeQA answers) from the model weights.

Across the three books, our performance is much higher than the \textit{no context} baseline. For \textit{full context}, we achieve three percent improvement in both BERTScore and Rouge F1 on Peter Rabbit, from 59 to 62 and 32 to 34. For Phantom of the Opera and Robinson Crusoe, \textsc{GroundedKG-RAG}'s Rouge-L F1 is relatively comparable to the full content with only 1/3 of its total length. The BERTScore drops from 62 to 56 and 55 to 46 compared to the full context, showing that longer books are still more challenging for the much bigger knowledge graph we built for these books.
Compared to GraphRAG we achieve similar performance on BERTScore, 62 compared to 63, but better performance on ROUGE-L, 34 compare to 8, indicating the difference in groundedness of the answers.
Due to the high number of LLM calls performed by GraphRAG, we limit our evaluation of the model to experiments on Peter Rabbit.

\textbf{Sentence Parsing.}
Comparing knowledge graph construction from SRL parses with AMR parses, we see in Table~\ref{tab:ablation-pr} that AMR parses improve on SRL parses in both BERTScore and ROUGE-L (52 vs. 58 and 30 vs. 34). This may be the result of the finer granularity level allowing more accurate node matches between query nodes and \textsc{GroundedKG} nodes. For example, in SRL parsing nodes, every token contributes equally to the node embedding, thus \textit{Peter's father} may have a higher cosine similarity score with \textit{Peter's mother} than \textit{the four little kids' father}. In AMR parsing, the core tokens determine the node embedding, thus the embedding of \textit{father} for both  \textit{Peter's father} and \textit{the four little kid's father} will lead to a match.

\textbf{TopK.}
We test the choice of TopK. $K=5$ receives lower BERTScore than $K=10$. The ROUGE-L is the same in Table~\ref{tab:ablation-pr}. Although the true grounded sentence has a higher probability in the first several nodes, we assume selecting more nodes allows more grounded sentences to be incorporated in the context, leading to better performance in the rare case that the gold answer is a reflection of multiple sentences. For instance, the node \textit{Peter} can be matched to a similar node \textit{Peter's jacket}, which is not in the very top of the matched nodes when compared to \textit{Peter}, \textit{Peter's sister} etc., yet contains the correct answer to the question.

\textbf{Embedding.}
We ablate the performance of our three embedding techniques on Peter Rabbit: basic node embedding, average neighbor embedding, and attention-based neighbor embedding. The three settings achieve similar scores on BERTScore. The original node embedding achieves the best score in BERTScore with 62 compared to 60 for others. The attention-based neighbor embedding gets the best score in ROUGE-L with 36. The average neighbor embedding gets the lowest scores. This is because the outliers in an average of embeddings can have a big influence on the node embedding, modifying up to 50\% of the top-10 retrieval results. The attention-based neighbor embedding is more stable compared to the average neighbor embedding and only changes 10-20\% of the retrieved results (often in the last ranks). Since, the contextual embedding, which considers the graph structure and the neighbors of a node does not exhibit an obvious gain on the task, we use the basic node embedding as our primary setting.

\textbf{Retrieval.}
We consider three different settings in the retrieval stage: basic node grounded\_text retrieval, filtering with sentence vector cosine similarity, and filtering with sentence retrieval counts. For sentence cosine similarity, the peformance drops 4 percent on BERTScore and 1 percent on ROUGE-L compared with grounded\_text retrieval for \textsc{GroundedKG-RAG}, and 1 percent in BERTScore and 5 percent in ROUGE-L when using SRL parses and $K=5$ on Peter Rabbit (see Table~\ref{tab:ablation-pr}). For filtering based on sentence retrieval counts, the BERTScore drops from 46 to 43 and F1 drops from 26 to 23 on Robinson Crusoe (see Table~\ref{tab:ablation-rc} in the appendix).
We conjecture that filtering can remove the gold grounded sentences for some of the questions, leading to worse question answering performance. During error analysis, we find that for the question \textit{``What did Peter do after arriving home?''}, the sentence \textit{``Peter arrives home exhaustedly.''} will have a high similarity score, while the true grounded text \textit{``Peter gets some food and runs to bed because he is too exhausted.''} receives a similarity score below the threshold because of the smaller number of matching tokens between the sentence and the question. The same happens for the retrieval count filter.

\section{Error Analysis}
We distinguish between (1) KG construction, (2) retrieval and (3) answer generation errors. The discussion of mapping \textit{``the four little kids' father''} to the father concept is an example of an error of the first kind (see Section~\ref{sec:results}). This works better in graphs built from AMR parses than from SRL parses. The second kind is a failure to retrieve relevant nodes from a given query (see an example in Table~\ref{tab:error-analysis} in the appendix). The third kind of error occurs when the answer generation model gets distracted by seemingly relevant information in the context, cannot handle the length of the context, or ignores the context and therefore the instruction (see third error type in Table~\ref{tab:error-analysis}).

\section{Graph Case Study}

We present a simple example to illustrate the creation of our \textsc{GroundedKG} from SRL and AMR parses in Table~\ref{tab:example-graphs}.
We can clearly observe that AMR-based graph grounded coreference-resolved entities (``it'' has been resolved to ``some camomile tea'' in the second clause) into the same node, in this case \textit{tea}, while SRL created a duplicate node for it. The same happens with the two mentions of Peter (``Peter'' and ``to Peter''). The AMR graph avoids duplicate node creation and improves node matching during retrieval.

\begin{table*}[ht]
\centering
\small 
\begin{tabular}{p{0.5\textwidth} p{0.5\textwidth}}
 \toprule
    \multicolumn{2}{p{\textwidth}}{%
    \textbf{Input Text:}
    \texttt{Peter's mother put Peter to bed and made some camomile tea;
    and Peter's mother gave a dose of some camomile tea to Peter!}
    }\\ \midrule
    AMR Graph: & SRL Graph: \\
    \begin{minipage}[t]{0.5\textwidth}
    \begin{verbatim}
(a / and
    :op1 (p / put-01
        :ARG0 (p2 / person
            :ARG0-of (h / have-rel-role-91
                :ARG1 (p3 / person
                    :name (n / name
                        :op1 "Peter"))
                :ARG2 (m / mother)))
        :ARG1 p3
        :ARG2 (b / bed))
    :op2 (m2 / make-01
        :ARG0 p2
        :ARG1 (t / tea
            :mod (c / camomile)
            :quant (s / some)))
    :op3 (d / dose-01
        :ARG0 p2
        :ARG1 p3
        :ARG2 t))
    \end{verbatim} 
    \end{minipage} & 
    \begin{minipage}[t]{0.5\textwidth}
    \begin{verbatim}
Predicate: put
    ARG-0: Peter's mother 
    ARG-1: Peter
    ARG-2: to bed 
    
Predicate: made
    ARG-0: Peter's mother
    ARG-1: some camomile tea
    
Predicate: gave
    ARG-0: Peter's mother
    ARG-1: a dose of some camomile tea
    ARG-2: to Peter 
    \end{verbatim}
    \end{minipage}\\ \midrule
    
    \textsc{GroundedKG} with AMR: & \textsc{GroundedKG} with SRL:\\

    \includegraphics[width=1\linewidth]{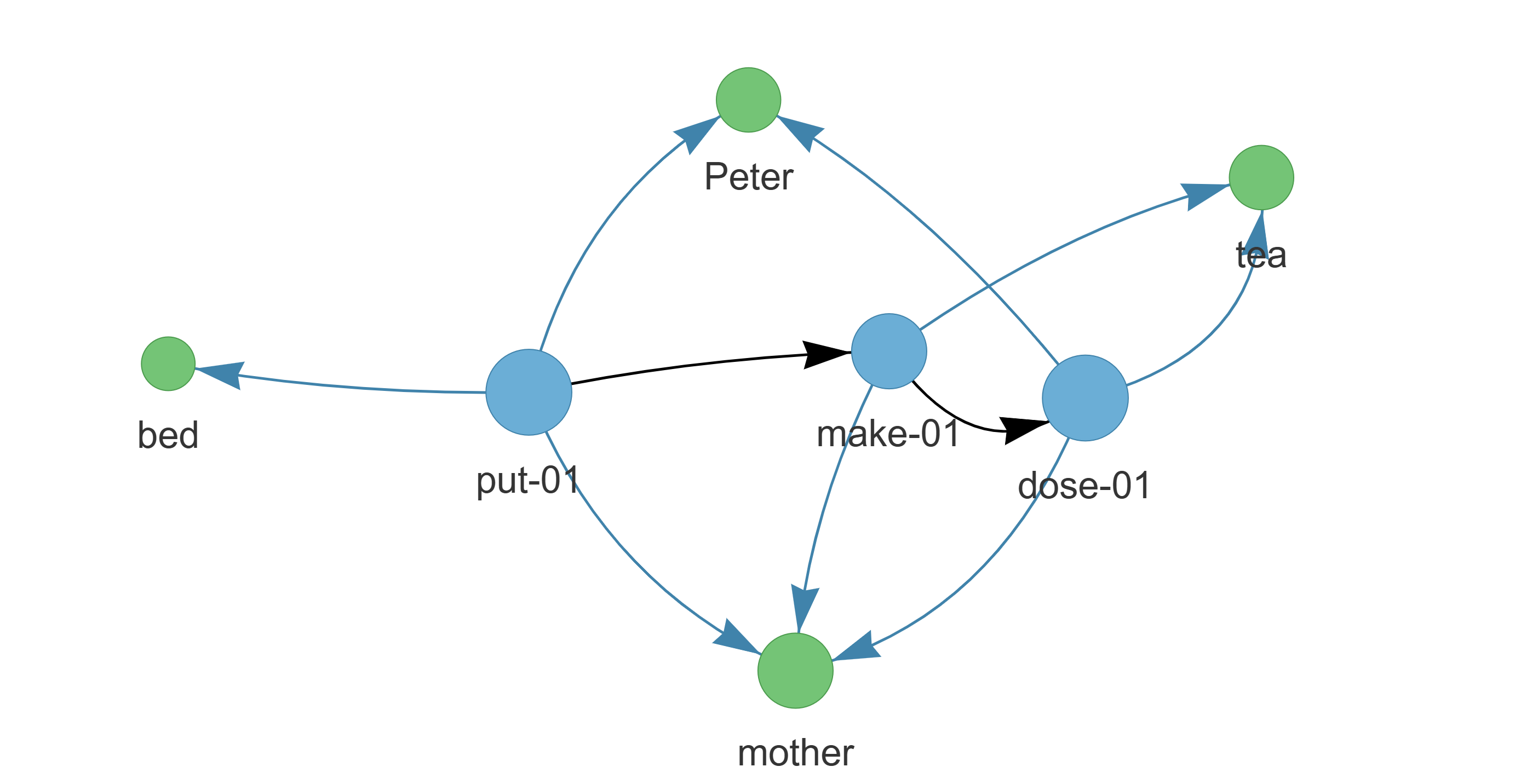} & \includegraphics[width=.7\linewidth]{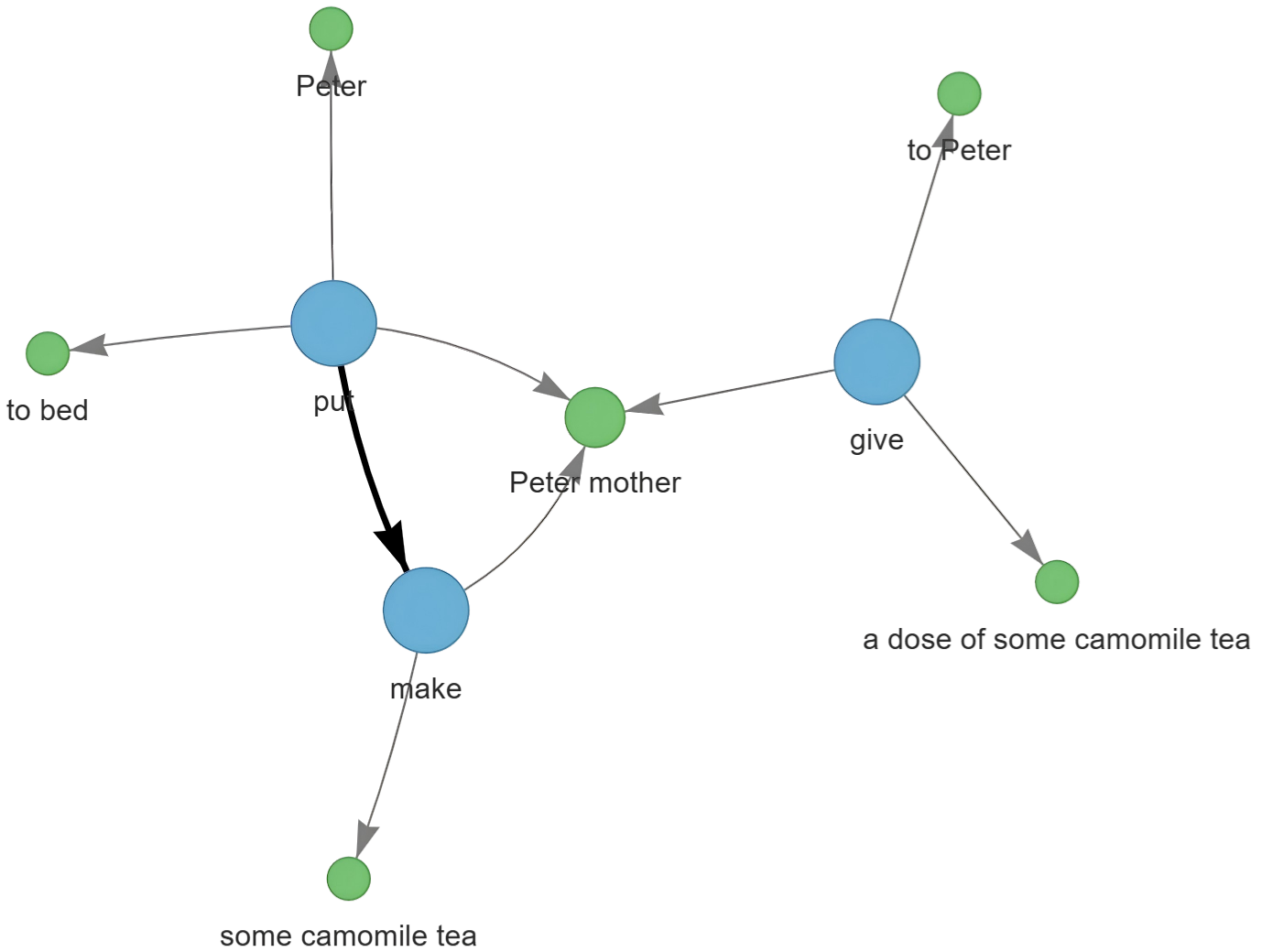} \\
    \bottomrule
\end{tabular}
\caption{Example comparing AMR and SRL graphs.}
\label{tab:example-graphs}
\end{table*}



\section{Conclusion}
In this work, we propose the \textsc{GroundedKG-RAG} framework, where each node and edge in the \textsc{GroundedKG} is directly extracted from and grounded in the source document at the sentence level. The nodes represent the entities and actions, and the edges represent the temporal or semantic relation between them. For the \textsc{GroundedKG} construction, we experimented with two different document parsing methods: SRL and AMR. Due to better semantic concept identification in AMR, it achieves higher scores on NarrativeQA than SRL. In our \textsc{GroundedKG}, we embed the nodes to incorporate the contexts. The graph's edges serve to relate nodes' temporal and semantic relationship. We evaluate three types of embedding techniques, and find that including neighboring nodes' information in a node's embedding does not significantly improve retrieval results. 
We also experiment with filtering retrieved grounded texts through either cosine similarity with the query embedding or retrieval counts, but neither filtering improves evaluation scores. Error analysis shows that these filters remove the gold sentence in a few evaluation examples.

In future work, we plan to further improve our \textsc{GroundedKG-RAG} in the following directions: (1) there is still room for improvement in the node matching, so different approaches to the graph structure, node/edge composition, used relations, and embedding techniques can be tried. (2) Even though we were not successful with our experiments in filtering out irrelevant source sentences, we believe that the answer generation model would profit from a less fragmented and sometimes contradicting context, giving an opportunity for better filtering techniques. (3) \textsc{GroundedKG-RAG}'s scores improving over the \textit{full context} baseline for the short Tale of Peter Rabbit, but BertScore is lagging behind on longer books shows that the large knowledge graphs constructed for these books may benefit from better graph retrieval techniques.

\section{Acknowledgements}
We thank getAbstract for the support and partial funding of this project.

\section{Bibliographical References}
\label{sec:reference}

\bibliographystyle{lrec2026-natbib}
\bibliography{references}

\appendix
\section{Question Answering Prompt}
We used the following prompt format to answer questions with and without context.

\label{app:rag-prompt}
\textbf{With content}: 
\texttt{Please answer the question based on the following content:}

\texttt{Content: $<content>$}

\texttt{Question: $<question>$}

\texttt{Answer:}

\textbf{Without content}: 
\texttt{Please answer the question based on your memory of the book $<book\_name>$:}

\texttt{Question: $<question>$}

\texttt{Answer:}

\section{Sentence Count Ablation}
A small ablation on Robinson Crusoe for the text filter with retrieval count is shown in Table~\ref{tab:ablation-rc}.

\begin{table*}
    \centering
    \small
    \begin{tabular}{ccccc}
    \toprule
Model & Exact match
& Sequence match& Bertscore& RougeL F1\\ \midrule
\multicolumn{5}{c}{Robinson Crusoe}\\\cmidrule(lr){1-5}
 No context
& 13
& 13
& 25
& 8
\\
 Full context
& 36
& 36
& \textbf{55}
& \textbf{31}
\\
 \textsc{GroundedKG-RAG}
& \textbf{40}
& \textbf{43}
& 46
& 30
\\
\textsc{GroundedKG-RAG} with RetCount $\ge 2$
& 30
& 30
& 43
& 23
\\  
\bottomrule 
    \end{tabular}%
    \caption{Comparing AMR with sentence count of 2 or more on Robinson Crusoe.}
    \label{tab:ablation-rc}
\end{table*}

\section{Error Analysis}
An error analysis with examples for the three types of errors we encounter is shown in Table~\ref{tab:error-analysis}.

\begin{table*}[ht]
\centering
\small
\begin{tabularx}{\textwidth}{l X}
 \toprule
    Tyep  I Error &\\  
    Question & What happened to the young rabbits' father?\\
    Matched Node & \makecell[X]{to the young rabbits' father ['with four little rabbits mother', 'The Tale Of Peter Rabbit', 'from Peter cousin, little Benjamin Bunny']\\
    happen ['happen\_112', 'come\_100', 'come\_113']} \\
    Grounded Text & text\_0-3, text\_0-6, text\_0-20, text\_0-21, text\_0-27, text\_0-37, text\_0-40, text\_0-42\\
    Predicted Answer & The text does not mention the father of the young rabbits.\\
    Gold Answer&Mrs. McGregor baked him into a pie.\\
    Solution& Improve \textsc{GroundedKG} construction for node matching\\ \midrule

    Type II Error &\\ 
    Question & What does Peter have for dinner after getting back home?\\
    Matched Node & \makecell[X]{Peter ['Peter', 'Peter back', 'First Peter']\\
    for dinner: [for supper, eat\_26, eat\_27]\\
    have: [have\_142, not have\_66, get\_131]\\
    Peter: [Peter, Peter back, First Peter]\\
    back home: [home, away, go back\_109]\\
    get: [get\_131, not get\_18, give\_2]}\\
    Grounded Text & \makecell[X]{text\_0-1, text\_0-8, text\_0-13, text\_0-14, text\_0-15, text\_0-16, text\_0-17, text\_0-18, text\_0-20, text\_0-21, text\_0-23, text\_0-24, text\_0-25, text\_0-27, text\_0-28, text\_0-29, text\_0-30, text\_0-31, text\_0-33, text\_0-35, text\_0-36, text\_0-37, text\_0-39, text\_0-40, text\_0-41, text\_0-42, text\_0-43, text\_0-44, text\_1-0, text\_1-1, text\_1-2, text\_1-4, text\_1-5, text\_1-7, text\_1-9, text\_1-11\\
    text\_1-9	His mother put him to bed and made some camomile tea;; and she gave a dose of some camomile tea to Peter!\\	
    text\_1-11	But-- Flopsy, Mopsy and Cottontail had bread and milk and blackberries for supper.}\\
    Predicted Answer & Bread and milk and blackberries.\\
    Gold Answer & Chamomile tea\\
    Solution & Filter out distracting results and use a stronger question answering model to distinguish the correct answer from misleading ones.\\ \midrule

    Type III Error\\
    Question&  Where does Peter see his lost clothing?\\
    Matched Node & \makecell[X]{Peter ['Peter', 'Peter back', 'First Peter']\\
    see ['see\_116', 'look\_30', 'look\_71']\\
    clothing ['his jacket', 'the little jacket and the shoes', 'underneath']\\
    lose ['lose\_41', 'lose\_42', 'lose\_136']}\\
    Grounded Text& \makecell[X]{text\_0-13, text\_0-14, text\_0-15, text\_0-16, text\_0-17, text\_0-18, text\_0-19, text\_0-20, text\_0-21, text\_0-23, text\_0-24, text\_0-25, text\_0-26, text\_0-27, text\_0-28, text\_0-29, text\_0-30, text\_0-31, text\_0-33, text\_0-35, text\_0-36, text\_0-37, text\_0-39, text\_0-40, text\_0-41, text\_0-42, text\_0-43, text\_0-44, text\_1-0, text\_1-1, text\_1-2, text\_1-3, text\_1-4, text\_1-5, text\_1-7, text\_1-9\\
    text\_1-3	Mr. McGregor hung up the little jacket and the shoes for a scare-crow to frighten the blackbirds.}\\
    Predicted Answer & In the tool-shed.\\
    Gold Answer & On McGregor's scarecrow\\
    Solution & Better filtering and using a question answering model that can reason over long and similar contexts. \\
    
\bottomrule
\end{tabularx}
\caption{Error analysis with an example for each type of error.}
\label{tab:error-analysis}
\end{table*}

\end{document}